\documentclass[11pt]{article}

\usepackage[utf8]{inputenc}
\usepackage[T1]{fontenc}
\usepackage{lmodern}
\usepackage[margin=1in]{geometry}
\usepackage[round,authoryear]{natbib}
\usepackage{hyperref}
\usepackage{url}
\usepackage{booktabs}
\usepackage{amsfonts}
\usepackage{amsmath}
\usepackage{amssymb}
\usepackage{nicefrac}
\usepackage{microtype}
\usepackage{xcolor}
\usepackage{array}
\usepackage{graphicx}
\usepackage{xspace}
\usepackage{algorithm}
\usepackage{algpseudocode}
\usepackage{subcaption}
\usepackage{placeins}

\hypersetup{hidelinks}
\makeatletter
\def\ps@preprintfirst{%
  \let\@mkboth\@gobbletwo
  \def\@oddhead{}%
  \def\@evenhead{}%
  \def\@oddfoot{\footnotesize Preprint.\hfil\normalfont\thepage\hfil}%
  \def\@evenfoot{\footnotesize Preprint.\hfil\normalfont\thepage\hfil}%
}
\makeatother

\graphicspath{{fig/}}

\newcommand{\OURS}{RoHIL\xspace}
\newcommand{\IRR}{\textsc{IRR}\xspace}
\newcommand{\Lfeat}{\mathcal{L}_{\mathrm{feat}}}
\newcommand{\Lmse}{\mathcal{L}_{\mathrm{mse}}}
\newcommand{\Lkl}{\mathcal{L}_{\mathrm{KL}}}
\newcommand{\Lbellman}{\mathcal{L}_{\mathrm{Bellman}}}
\newcommand{\Lcritic}{\mathcal{L}_{\mathrm{Critic}}}
\newcommand{\Lactor}{\mathcal{L}_{\mathrm{Actor}}}
\newcommand{\Lsac}{\mathcal{L}_{\mathrm{SAC}}}

\newcommand{\sg}{\mathrm{sg}}

\title{\textbf{\OURS: Robust Human-in-the-Loop Robotic Reinforcement Learning Against Illumination Variations}}

\author{%
\textbf{Shuoqin Zhang}$^{1,2}$ \quad
\textbf{Yixin Xiong}$^{1}$ \quad
\textbf{Xiru Gao}$^{1}$ \quad
\textbf{Kai Liu}$^{1}$\\
\textbf{Ke Wang}$^{1}$ \quad
\textbf{Xichuan Zhou}$^{1,\dagger}$ \quad
\textbf{Zhe Hu}$^{1,2,\dagger}$\\[0.5em]
$^{1}$Chongqing University \quad
$^{2}$Chengdu Anu Intelligence\\[0.25em]
$^{\dagger}$Corresponding Author%
}
\date{}

\begin{document}

\maketitle
\thispagestyle{preprintfirst}

\begin{abstract}
Human-in-the-loop reinforcement learning systems achieve near-perfect success on the workstation where they are trained, but collapse when the same robot is moved to a workstation a few meters away due to shifts in the visual input distribution caused by new lamp positions and window light. Re-collecting demonstrations and re-running HIL on every workstation is incompatible with deployment, and naively fine-tuning on shifted-light data triggers catastrophic forgetting of the source workstation. To close this cross-domain gap, we present \OURS, an offline fine-tuning framework that uses no extra real-robot interaction. \OURS combines (i)~a \emph{world-model-based image relighter} that re-synthesises the visual stream of source-workstation trajectories under multiple virtual HDRI environments, leaving actions and rewards real; (ii)~\emph{Illumination-Retention Replay} (\IRR), a data-level anti-forgetting mechanism that interleaves relit adaptation transitions with original-light retention transitions to preserve source-workstation Bellman coverage; and (iii)~an \emph{anchored Bellman--actor regulariser} that constrains representation and policy drift from the original source-workstation policy. Across four real-robot manipulation tasks under significant cross-workstation illumination variations, \OURS substantially improves shifted-light performance where standard HIL-RL collapses, while preserving source-workstation performance, eliminating the need to re-collect data and retrain for every new workstation and environment. Project page: \url{https://anonymous4365.github.io/RoHIL/}.
\end{abstract}

\section{Introduction}
\label{sec:intro}

Human-in-the-loop reinforcement learning (HIL-RL), exemplified by HIL-SERL~\citep{luo2025hilserl}, has rapidly become the de facto recipe for sample-efficient real-world robotic manipulation: a few hundred human demonstrations seed a SAC~\citep{haarnoja2018sac,ball2023rlpd} learner, an operator intervenes to correct failure modes, and within a few hours of online interaction the policy reaches near-perfect success on contact-rich tasks. However, existing HIL-RL methods still degrade severely under illumination variation between training and deployment, and the most common remedies (pixel jitter, demo re-collection, naive fine-tuning) either fail to reproduce real light transport or destroy the source-workstation policy.

\begin{figure}[t]
    \centering
    \includegraphics[width=0.95\linewidth]{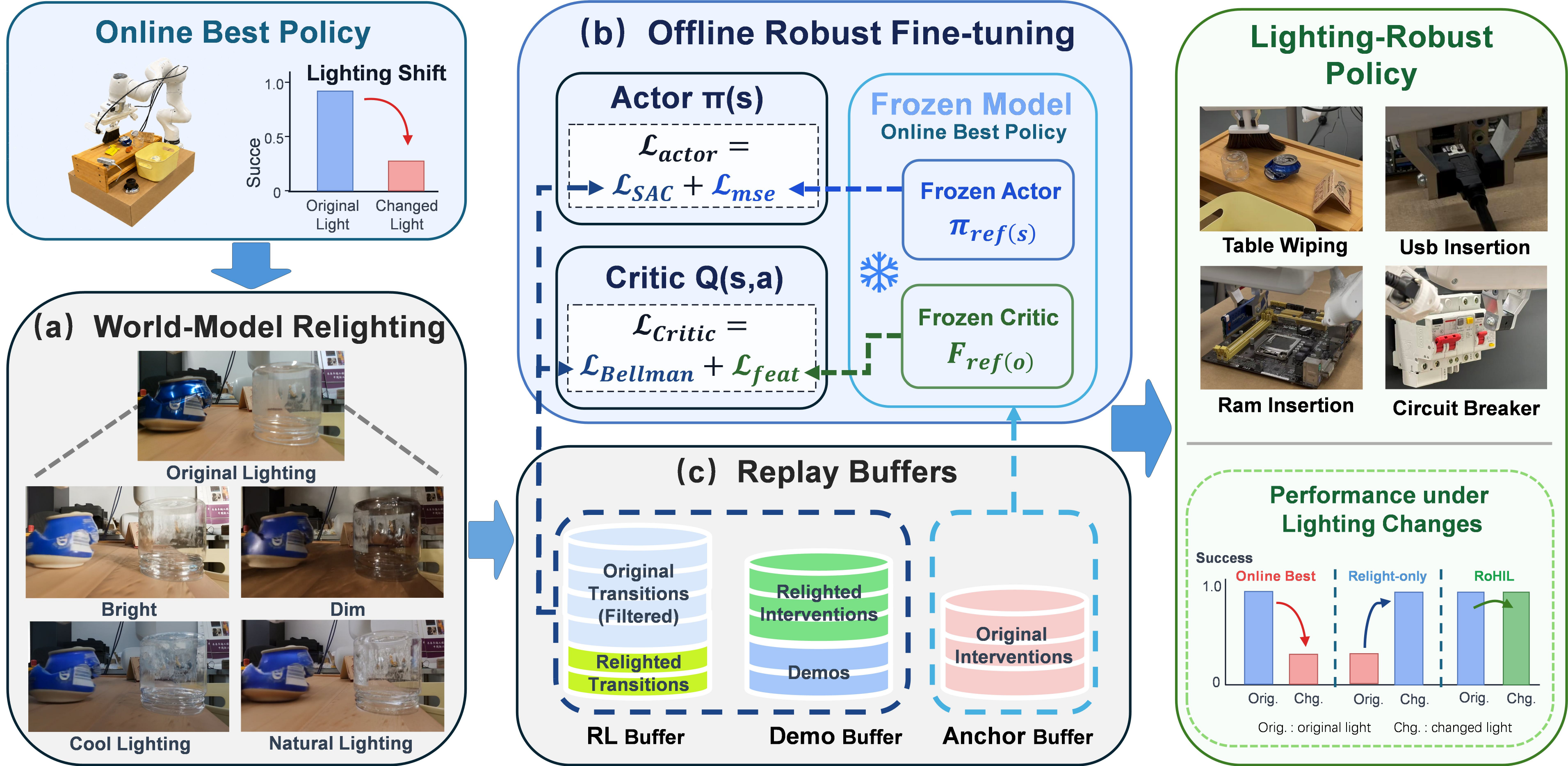}
    \caption{\textbf{\OURS overview.} Starting from an online best policy that degrades under lighting shifts, \OURS uses (a) world-model relighting to synthesise illumination-diverse transitions, (b) offline robust fine-tuning with $\Lactor{=}\Lsac{+}\Lmse$ and $\Lcritic{=}\Lbellman{+}\Lfeat$, and (c) \IRR replay buffers for retention. The resulting policy preserves original-light performance while improving changed-light robustness across four real-robot tasks.}
    \label{fig:overview}
\end{figure}

\paragraph{The hidden ceiling of HIL-RL: lack of robustness under domain shift.}
HIL-RL trains a policy within a closed source visual domain, but deployment exposes the same policy to many target domains induced by changes in lamp placement, daylight direction, shadows, and specular highlights. Since these illumination domains are not known in advance, a naive deployment recipe must treat each workstation as a new training problem: collect new demonstrations and run a new HIL session. The human-interaction cost therefore scales as $O(N)$ in the number of workstations. Our goal is to make this cost $O(1)$: train once on a source workstation, then adapt the resulting policy offline to unseen illumination domains without additional robot interaction. \OURS realizes this goal by relighting the source trajectories and fine-tuning with retention and anchoring, allowing a single HIL run to be amortized across deployment-time lighting environments.

\paragraph{Why is this hard?}
Cross-workstation lighting shift is difficult because the natural remedies are either insufficient or incompatible with deployment. Simple pixel-level augmentation can perturb image appearance, but it does not capture the structured illumination changes that arise from different lamps, daylight directions, shadows, and specular highlights. Collecting new demonstrations and re-running HIL on each workstation would capture these effects, but it restores the $O(N)$ deployment cost that we aim to avoid. Offline fine-tuning offers a scalable alternative, but naively adapting the source-workstation policy to shifted-light data can overwrite the representation and policy that solved the original workstation, causing catastrophic forgetting.

These failures leave three problems that must be solved:
(i)~obtaining realistic illumination variation without new robot interaction;
(ii)~using this variation to adapt the policy to shifted lighting domains; 
(iii)~preserving the source-workstation policy during adaptation.
\OURS addresses these problems with world-model relighting, Illumination-Retention Replay, and an anchored Bellman--actor regulariser.

\paragraph{Our approach.}
We propose \OURS, an offline fine-tuning framework that adapts a source-workstation HIL-SERL policy to illumination shifts without new robot interaction. \OURS relights recorded trajectories to expose the policy to new lighting, uses \IRR as a data-level retention mechanism during replay, and anchors the Bellman and actor updates to a frozen source policy. Together, these components enable robust illumination generalization.

\paragraph{Contributions.}
The combination of these components, summarized in Figure~\ref{fig:overview}, defines \OURS:
\begin{itemize}
    \item We formulate cross-workstation HIL-RL deployment under illumination shift as an offline domain-adaptation problem, and propose \OURS, a framework that amortizes one source-workstation HIL-SERL run into an $O(1)$ deployment procedure with no additional real-robot interaction.

    \item We introduce a relight-and-retain data pipeline for robust adaptation. A world-model relighter generates illumination-diverse observations from already-collected trajectories while preserving real actions and rewards, and \IRR preserves source-workstation Bellman coverage by balancing relit adaptation data with original-light retention data.

    \item We design an anchored Bellman--actor regulariser for safe fine-tuning under lighting shift. The regulariser adds $\Lfeat$ to the critic update and $\Lmse$ to the actor update, mitigating source-domain forgetting at both the representation and policy levels while retaining plasticity to relit observations.
\end{itemize}

\section{Related Work}
\label{sec:related}

\paragraph{Vision-based robotic RL with humans in the loop.}
HIL-SERL~\citep{luo2025hilserl} combines SAC~\citep{haarnoja2018sac}, RLPD-style symmetric replay~\citep{ball2023rlpd}, and SpaceMouse interventions to solve contact-rich tasks from modest demonstrations. Related HIL/RL and imitation baselines include IBRL~\citep{hu2024ibrl}, ResiP~\citep{ankile2024resip}, DAgger / HG-DAgger~\citep{ross2011dagger,kelly2019hgdagger}, behaviour cloning, ACT~\citep{zhao2023act}, and Diffusion Policy~\citep{chi2025diffusion}. Pixel augmentations such as RAD~\citep{laskin2020rad} and DrQ~\citep{kostrikov2021drq,yarats2022drqv2} improve sample efficiency, but their low-order image perturbations do not model cross-workstation light transport; \OURS instead performs offline relighting and retention-aware fine-tuning.

\paragraph{World models for image relighting.}
Recent video relighters provide a stronger source of illumination variation than hand-designed image noise. On the Cosmos-Transfer1 video-diffusion backbone~\citep{alhaija2025cosmos,blattmann2023svd}, DiffusionRenderer~\citep{liang2025diffusionrenderer} performs HDRI-conditioned inverse and forward rendering, while UniRelight~\citep{he2025unirelight} improves temporal consistency through joint decomposition--synthesis. We adopt DiffusionRenderer for its resource-efficiency trade-off (\S\ref{sec:method-relighting}, Appendix~\ref{app:relighting-compare}); other candidates including IC-Light~\citep{zhang2024iclight}, Lumos~\citep{liu2025lumoscustom}, Lumen~\citep{yang2025lumen}, TC-Light~\citep{liu2025tclight}, and VidToMe~\citep{li2024vidtome} were less reliable on indoor manipulation views. Prior robot-lighting robustness mostly relies on domain randomisation or saliency-guided augmentation~\citep{tobin2017domain,zhuang2025robosaga,jin2025pblighting}.

\paragraph{Fine-tuning without forgetting.}
Offline-to-online RL methods such as CQL~\citep{kumar2020cql}, IQL~\citep{kostrikov2022iql}, Cal-QL~\citep{nakamoto2023calql}, RLDG~\citep{xu2025rldg}, ConRFT~\citep{chen2025conrft}, DPPO~\citep{ren2025dppo}, and WSRL~\citep{zhou2025wsrl} study adaptation from static data, while generalist VLA models~\citep{brohan2023rt1,brohan2023rt2,kim2024openvla,octo2024,black2024pi0,open_x_embodiment} address broader skill transfer. Our setting is narrower but sharper: adapting a solved HIL policy to lighting shift without regressing on the source workstation. \IRR is inspired by continual replay, where retaining the historical transition distribution prevents value regression~\citep{rolnick2019clear}; our anchored Bellman--actor regularizer follows frozen-reference learning without forgetting and policy distillation~\citep{li2018lwf,rusu2016policy}, with high-quality expert anchors motivated by exemplar-memory methods~\citep{rebuffi2017icarl,lopezpaz2017gem,buzzega2020derpp}.

\section{Preliminaries and Problem Setup}
\label{sec:prelim}

\paragraph{HIL-SERL with RLPD-style symmetric replay.}
We start from the HIL-SERL training stack~\citep{luo2025hilserl}: a SAC actor--critic~\citep{haarnoja2018sac} with a ResNet-10 visual encoder $\phi_\theta(o)$ ingests wrist-camera RGB and proprioception. The policy $\pi_\theta(a\mid s)$ is a \emph{continuous} squashed (tanh) diagonal-Gaussian over a 6-dimensional end-effector delta twist~\citep{haarnoja2018sac}, with mean $\mu_\theta(s)$ and diagonal log-variance $\log\sigma_\theta^2(s)$ produced by the actor head. An RLPD-style buffer~\citep{ball2023rlpd} draws each minibatch with a $50/50$ symmetric split between an online RL replay $\mathcal{R}$ and a demonstration buffer $\mathcal{D}$:
\[
\mathcal{B} \;=\; \tfrac{1}{2}\,\mathrm{Sample}(\mathcal{R}) \,\cup\, \tfrac{1}{2}\,\mathrm{Sample}(\mathcal{D}).
\]
The critic is trained on the joint Bellman loss $\Lbellman = \mathbb{E}_{\mathcal{B}}[(Q_\theta(s,a)-y)^2]$ with $y=r+\gamma(1-d)\,\mathbb{E}_{a'\sim\pi_\theta(\cdot|s')}[Q_{\bar\theta}(s',a')-\eta\log\pi_\theta(a'|s')]$; the actor on $\Lsac=\mathbb{E}_{s\sim\mathcal{B},\,a\sim\pi_\theta(\cdot|s)}[\eta\log\pi_\theta(a|s)-Q_\theta(s,a)]$, where $\eta$ is the SAC entropy temperature (we reserve $\alpha$ for the \IRR retention coefficient introduced in \S\ref{sec:method-buffer}). A SpaceMouse intervention channel injects expert corrections, which are streamed simultaneously into $\mathcal{R}$ (as transitions) and $\mathcal{D}$ (as the demonstration set).

\paragraph{Cross-workstation deployment as Domain-Incremental Learning.}
We define a \emph{workstation} as a fixed combination of robot, parts, and ambient illumination. At training time we observe the source workstation $W_0$; at deployment time we encounter a sibling workstation $W_k$ that shares the same task semantics, kinematics, contact dynamics, and reward function as $W_0$, but exhibits a shifted visual observation distribution $p_{W_k}(o\mid s)\neq p_{W_0}(o\mid s)$, induced by changes in window orientation, ceiling and task lighting, and ambient occlusion. The agent has no access to a workstation identifier at inference time, must perform on $W_k$, and, critically for deployment, must \emph{not regress} on $W_0$. This is precisely the Domain-Incremental Learning regime~\citep{rolnick2019clear,li2018lwf}.

\paragraph{Why naive fine-tuning fails: the stability/plasticity dilemma.}
A naive remedy, namely fine-tuning the source-workstation policy on data collected (or synthesised) under shifted lighting, creates a textbook stability/plasticity dilemma. Increasing plasticity via vanilla SGD on relit data drives the encoder $\phi_\theta$ toward the new visual statistics, but the same gradient updates erode source-workstation features, collapsing $W_0$ success well below the pre-fine-tune level. Conversely, overweighting original-light data starves the policy of new-domain signal. \OURS resolves this dilemma along two complementary anti-forgetting axes: (i)~\IRR, which balances original-light retention with relit adaptation to maintain Bellman coverage on $W_0$ (\S\ref{sec:method-buffer}), and (ii)~an anchored Bellman--actor regularizer that constrains encoder and policy outputs on source-workstation expert states to match the frozen pre-fine-tune model (\S\ref{sec:method-loss}).

\section{Method}
\label{sec:method}

\OURS is an \emph{offline} fine-tuning framework that turns a single source-workstation HIL-SERL run into a policy robust to cross-workstation illumination shift, with no further real-robot interaction required after Stage~1 (\S\ref{sec:method-relighting}). The design goal is to suppress source-domain forgetting during offline adaptation while still exposing the policy to visual evidence needed for lighting-invariant generalisation. It comprises three stages (Figure~\ref{fig:overview}): (\textbf{Stage~1}, \S\ref{sec:method-relighting}) world-model relighting supplies new illumination signals for adaptation while preserving real actions and rewards; (\textbf{Stage~2}, \S\ref{sec:method-buffer}) \IRR provides data-level anti-forgetting by retaining original-light Bellman coverage inside the RLPD replay structure; (\textbf{Stage~3}, \S\ref{sec:method-loss}) anchored Bellman--actor regularisation provides representation- and policy-level anti-forgetting through $\Lcritic{=}\Lbellman{+}\Lfeat$ and $\Lactor{=}\Lsac{+}\Lmse$. The three components are coupled by design: \IRR alone leaves the encoder free to drift, anchoring alone lacks Bellman coverage of source-workstation negative states, and only their combination closes the gap (\S\ref{sec:experiments}).

\subsection{Stage 1: Source-Workstation Collection and World-Model Relighting}
\label{sec:method-relighting}

A HIL-SERL session on the source workstation gives us real robot trajectories, but only under one illumination domain. Stage~1 (\S\ref{sec:method-relighting}) turns these source trajectories into shifted-light adaptation data without collecting new robot interaction. Each transition is stored as $(o_t, a_t, r_t, o_{t+1}, d_t)$, where the visual components of $o_t$ and $o_{t+1}$ are the parts affected by cross-workstation lighting changes. The action, reward, and termination label, however, come from the recorded physical interaction and should remain unchanged. \OURS therefore preserves $a_t$, $r_t$, and $d_t$, and re-synthesises the visual stream of each trajectory under multiple lighting conditions. This produces illumination-diverse observations while keeping the supervision signal grounded in real robot experience.

\paragraph{World-model relighter.}
We instantiate the relighting module with Cosmos-Transfer1-Diffusion\allowbreak{}Renderer~\citep{liang2025diffusionrenderer}, an HDRI-conditioned video relighter built on the NVIDIA Cosmos-Transfer1 video-diffusion foundation~\citep{alhaija2025cosmos}. Given a recorded RGB stream and a target HDRI environment map, the model produces a relit RGB stream that changes the illumination state while preserving the scene layout, object geometry, and material appearance. We synthesise four target lighting conditions for each recorded trajectory (Appendix~\ref{app:lighting-conditions}, Figure~\ref{fig:lighting-grid}). Together with the original lighting, this expands each source trajectory into five illumination variants that share the same actions, rewards, and termination labels.

\paragraph{Relighter implementation.}
We choose DiffusionRenderer for its quality--cost trade-off in indoor manipulation views. UniRelight~\citep{he2025unirelight}, another Cosmos-based relighter, gives slightly stronger temporal consistency but is substantially more expensive: relighting one camera stream of $8\,000$ transitions takes $6.82$~h and $26.80$~GB peak VRAM with DiffusionRenderer, versus $48.63$~h and $39.88$~GB with UniRelight (Appendix~\ref{app:relighting-compare}, Table~\ref{tab:relighter-cost}). Additional relighter pilots are reported in Appendix~\ref{app:relighting-compare}.

\subsection{Stage 2: Illumination-Retention Replay}
\label{sec:method-buffer}
\paragraph{Four data pools.}
Stage~1 (\S\ref{sec:method-relighting}) expands the source-workstation data into the four transition pools in
Table~\ref{tab:irr-pools}. The two policy-driven pools contain transitions whose
actions were produced by the trained policy, and therefore include both successful
and unsuccessful source-workstation experience. The two demonstration pools
contain human actions and provide higher-quality expert support. The original-light
and relit versions are disjoint visual domains, while sharing the same underlying
action and reward labels.

\begin{table}[t]
\centering
\footnotesize
\setlength{\tabcolsep}{6pt}
\caption{\textbf{Transition pools used by IRR.}
Superscript $0$ denotes original lighting, and superscript $\mathrm{rel}$ denotes relit observations.}
\label{tab:irr-pools}
\begin{tabular}{@{}lccc@{}}
\toprule
\textbf{Pool} & \textbf{Symbol} & \textbf{Light} & \textbf{Action} \\
\midrule
Policy-driven RL        & $\mathcal{R}^{0}_{\pi}$              & source & mixed  \\
Human demos             & $\mathcal{D}^{0}$                    & source & expert \\
Relit policy-driven RL  & $\mathcal{R}^{\mathrm{rel}}_{\pi}$   & relit  & mixed  \\
Relit demos             & $\mathcal{D}^{\mathrm{rel}}$         & relit  & expert \\
\bottomrule
\end{tabular}
\end{table}
\paragraph{From RLPD replay to \IRR.}
RLPD's symmetric replay assumes that the RL pool and demonstration pool share one observation domain. After relighting, naive RLPD faces a domain-allocation problem: relit-only replay maximises adaptation but deletes source-workstation negative states, while original-only replay preserves the source policy but starves the critic of new illumination evidence. Inspired by continual replay's retention principle~\citep{rolnick2019clear}, \IRR keeps the RLPD $50/50$ structure but inserts a retention coefficient inside the RL half:
\begin{equation}
    \begin{aligned}
    \mathcal{R}_\alpha &= \alpha\,\mathcal{R}^{0}_{\pi} + (1-\alpha)\bigl(\mathcal{R}^{\mathrm{rel}}_{\pi}\cup\mathcal{D}^{\mathrm{rel}}\bigr),\quad
    \mathcal{D}_{\mathrm{anc}} = \mathcal{D}^{0}\cup\mathcal{D}^{\mathrm{rel}},\\
    \mathcal{B} &= \tfrac{1}{2}\mathrm{Sample}(\mathcal{R}_\alpha)\cup\tfrac{1}{2}\mathrm{Sample}(\mathcal{D}_{\mathrm{anc}}).
    \end{aligned}
    \label{eq:irr}
\end{equation}
Here $\mathcal{R}^{0}_{\pi}$ is the original-light policy-driven pool, $\mathcal{R}^{\mathrm{rel}}_{\pi}$ its relit counterpart, and $\mathcal{D}^{0},\mathcal{D}^{\mathrm{rel}}$ are original-light and relit demonstrations. Thus \IRR is a data-level anti-forgetting mechanism: the relit component supplies illumination-adaptation signal, while the retained original-light component preserves Bellman coverage of $W_0$.

\paragraph{Why a strict majority of original-light data in $\mathcal{R}_\alpha$?}
The original-light policy-driven pool is the \emph{full distribution} of source-workstation experience, including successful trajectories, exploratory states, negative states, and occasional emergent behaviours beyond the human demonstrations. Discarding it removes the only source-workstation failures the critic has seen and encourages over-estimation of $Q$ on unseen $W_0$ states. We therefore sweep $\alpha\in\{0.0, 0.05, \ldots, 0.95, 1.0\}$ and find $\alpha=0.75$ to be the consistent best operating point on both source and shifted lighting (\S\ref{sec:exp-buffer-sweep}).

\paragraph{Why anchor only on expert data.}
If policy-driven transitions were also used as anchors, the Stage~3 (\S\ref{sec:method-loss}) anchor terms would pin the policy to match the frozen model on failure states, literally instructing it to ``replicate failures'' (the GEM~\citep{lopezpaz2017gem} / iCaRL~\citep{rebuffi2017icarl} failure mode for low-quality exemplars). \IRR instead uses policy-driven data only for unconstrained Bellman coverage, while $\mathcal{D}_{\mathrm{anc}}$ supplies high-quality expert support for Stage~3 (\S\ref{sec:method-loss}).

\subsection{Stage 3: Anchored Bellman--Actor Regularization}
\label{sec:method-loss}

\paragraph{Source-anchored fine-tuning.}
\IRR (\S\ref{sec:method-buffer}) preserves source-workstation data in replay, but replay alone does not constrain how the model changes during offline fine-tuning. Bellman updates on relit observations can still shift the visual encoder, alter critic targets, and drive the actor away from the solved source policy. Stage~3 (\S\ref{sec:method-loss}) therefore adds frozen-source anchors on expert states. The feature anchor preserves source-domain representations, while the actor anchor preserves reference action preferences. Together, they allow relit data to improve illumination robustness while suppressing source-domain forgetting.

\paragraph{SAC update under \IRR.}
Let $\theta_0$ denote the frozen source-workstation parameters, and let $\mathcal{B}$ and $\mathcal{D}_{\mathrm{anc}}$ be the \IRR minibatch and anchor pool from Equation~\ref{eq:irr}. For twin critics $Q_{\theta_i}$ and target critics $Q_{\bar\theta_i}$, the relit/original-light Bellman update is
\begin{align}
    y &= r+\gamma(1-d)\,\mathbb{E}_{a'\sim\pi_\theta(\cdot|s')}\!\left[\min_{i=1,2}Q_{\bar\theta_i}(s',a')-\eta\log\pi_\theta(a'|s')\right], \label{eq:sac-target}\\
    \Lbellman &= \mathbb{E}_{(s,a,r,s',d)\sim\mathcal{B}}\!\left[\sum_{i=1}^{2}\bigl(Q_{\theta_i}(s,a)-\sg(y)\bigr)^2\right], \label{eq:lbellman}\\
    \Lsac &= \mathbb{E}_{s\sim\mathcal{B},\,a\sim\pi_\theta(\cdot|s)}\!\left[\eta\log\pi_\theta(a|s)-\min_{i=1,2}Q_{\theta_i}(s,a)\right],
    \label{eq:lsac}
\end{align}
where $\eta$ is the SAC entropy temperature. These terms provide the plasticity needed to fit the relit observation distribution, but alone they do not prevent the source-domain representation from moving.

\paragraph{Anchors for representation and policy retention.}
We therefore add two frozen-reference anchors on $\mathcal{D}_{\mathrm{anc}}$. The feature anchor constrains the visual representation used by both critics and actor:
\begin{equation}
    \Lfeat = \lambda_{\mathrm{feat}}\,\rho(t)\,\mathbb{E}_{o\sim\mathcal{D}_{\mathrm{anc}}}\!\bigl[\|\phi_\theta(o)-\sg(\phi_{\theta_0}(o))\|_F^2\bigr].
    \label{eq:lfeat}
\end{equation}
The actor-side reference-action anchor is applied inside $\Lactor$: rather than treating the squared mean penalty as an isolated imitation loss, $\Lmse$ regularises the pre-tanh Gaussian action mean while SAC continues to optimise value-seeking behaviour:
\begin{equation}
    \Lmse = \beta_{\mathrm{mse}}\,\rho(t)\,\mathbb{E}_{s\sim\mathcal{D}_{\mathrm{anc}}}\!\bigl[\|\mu_\theta(s)-\sg(\mu_{\theta_0}(s))\|_2^2\bigr],
    \label{eq:lmse}
\end{equation}
where $\mu_\theta(s)$ is the diagonal-Gaussian mean. We set $\lambda_{\mathrm{feat}}=0.2$ and $\beta_{\mathrm{mse}}=0.1$.

\paragraph{Full anchored objective.}
The final fine-tuning step couples adaptation and retention as
\begin{equation}
    \Lcritic = \Lbellman + \Lfeat,\qquad
    \Lactor = \Lsac + \Lmse.
    \label{eq:anchored-objective}
\end{equation}
To avoid over-constraining late adaptation, both anchors decay with $\rho(t)=1-(1-\rho_{\mathrm{end}})t/T$ and $\rho_{\mathrm{end}}=0.33$, shifting the update from early retention to late illumination generalisation. Appendix~\ref{app:actor-head} further compares the reference-action mean anchor with a full Gaussian KL anchor and shows that KL is not the optimal choice in our setting.

\section{Experiments}
\label{sec:experiments}

Our empirical evaluation is organised around three studies, each isolating one piece of the design, with the last bringing the full system into contact with established baselines. We first establish, on the most lighting-sensitive task (USB insertion), that the anchored Bellman--actor objective is a strictly better fine-tune objective than the standard SAC loss and that its two terms contribute independently (\S\ref{sec:exp-loss-ablation}). We then sweep the original-vs.-relit mixing ratio in \IRR to identify the optimal retention balance and verify that the loss advantage holds across the full sweep (\S\ref{sec:exp-buffer-sweep}), and put \IRR and the anchored objective into a $2{\times}2$ ablation to disentangle their individual contributions (\S\ref{sec:exp-axis-ablation}). Finally we benchmark the full \OURS system against five HIL-compatible baselines on four real-robot tasks under ten illumination conditions (\S\ref{sec:exp-cross-method}).

\subsection{Tasks, evaluation protocol, and baselines}
\label{sec:exp-setup}

\begin{figure}[t]
    \centering
    \includegraphics[width=\linewidth]{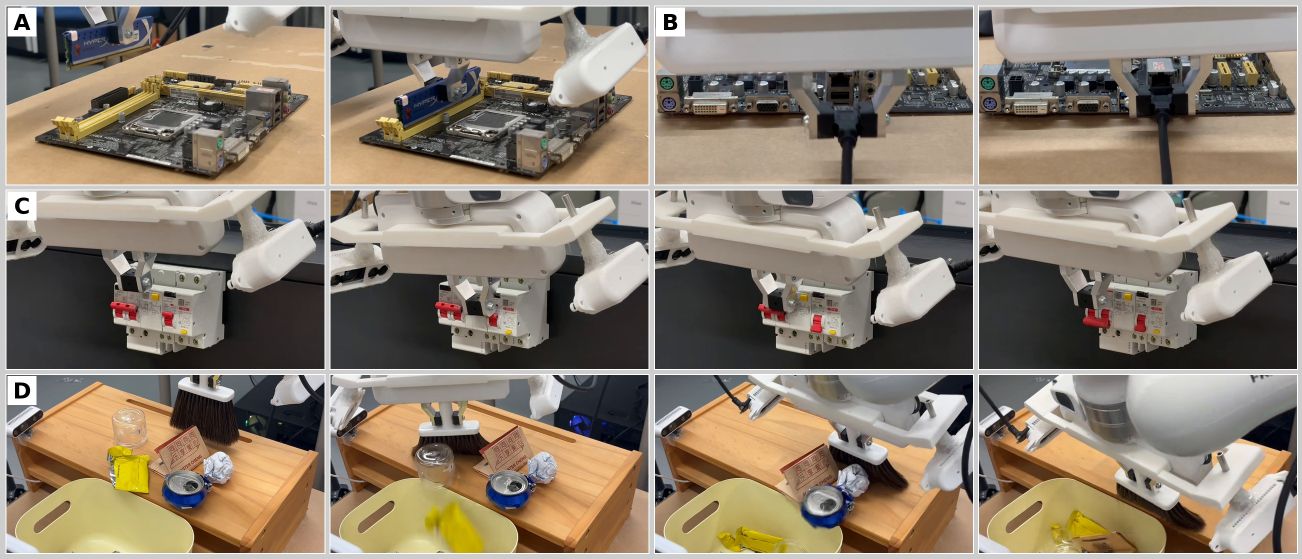}
    \caption{The four manipulation tasks: (a) \texttt{ram\_insertion}, (b) \texttt{usb\_insertion}, (c) \texttt{circuit\_breaker}, and (d) \texttt{table\_wiping}. Per-task hardware setup is in Appendix~\ref{app:tasks}.}
    \label{fig:tasks}
\end{figure}

We evaluate on four real-robot manipulation tasks executed on a Franka Emika Panda arm with a wrist-mounted RealSense camera (Figure~\ref{fig:tasks}): \texttt{ram\_insertion}, \texttt{usb\_insertion}, \texttt{circuit\_breaker}, and \texttt{table\_wiping}. Each task is collected on a designated source workstation. For deployment-time evaluation, we sweep an illumination-shift intensity gradient at $\{0\%,10\%,\ldots,90\%,100\%\}$ deviation from the source-workstation baseline; the $60\%$-shift point is used as the headline cross-workstation evaluation. Lighting conditions cover five HDRI maps, three task-light spotlight configurations, and two natural-window-light shifts (Appendix~\ref{app:lighting-conditions}). Source training uses task-dependent HIL-SERL budgets (RAM $60$k, USB $30$k, circuit-breaker $35$k, table wiping $95$k steps); \OURS fine-tunes offline for $15\,000$ steps with no further real-robot interaction. Per-task hyperparameters are deferred to Appendix~\ref{app:tasks}; results below report $30$ trajectories per condition, except for the cross-method headline which uses $60$.

We compare \OURS against five HIL-compatible baselines, all retrained on the same source-workstation data: HIL-SERL~\citep{luo2025hilserl}, HG-DAgger~\citep{kelly2019hgdagger}, behaviour cloning (BC), IBRL~\citep{hu2024ibrl}, and ACT~\citep{zhao2023act}. The other relighter candidates we surveyed (IC-Light, Lumos, Lumen, TC-Light, VidToMe) are discussed but not benchmarked because they failed pilot reproductions on indoor manipulation views (\S\ref{sec:related}).

\subsection{Loss study: anchored objective outperforms standard SAC}
\label{sec:exp-loss-ablation}

\begin{figure}[t]
    \centering
    \includegraphics[width=0.78\linewidth]{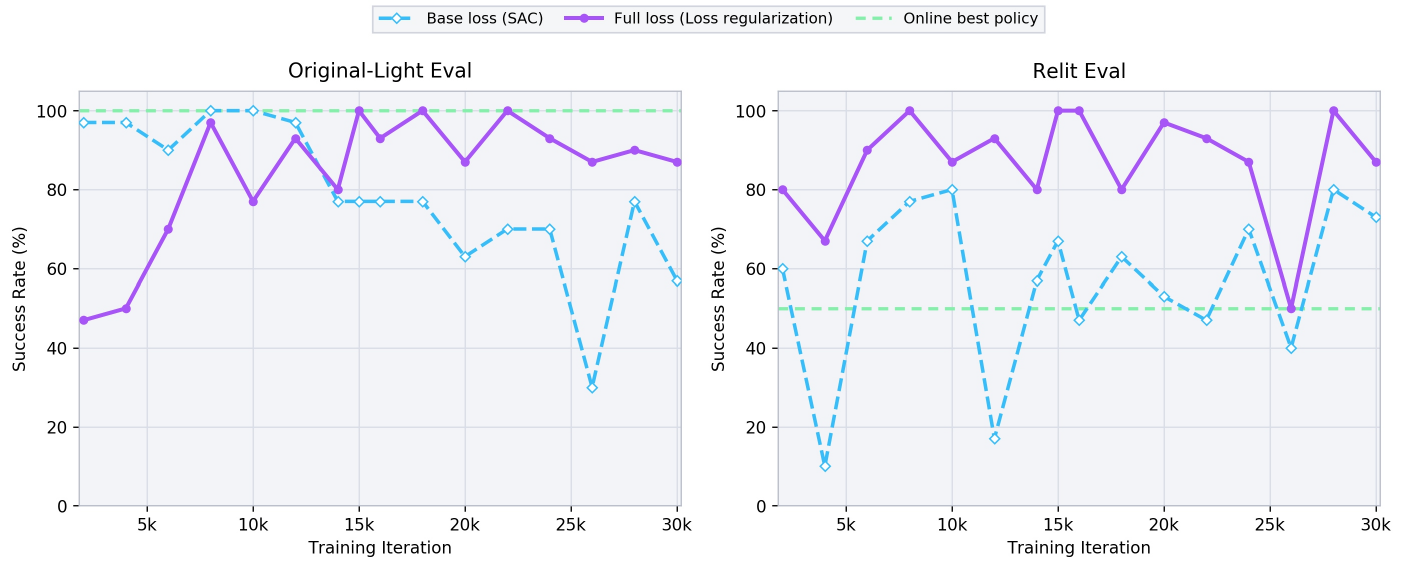}
    \caption{USB insertion ($\alpha{=}0.75$): training-iteration sweep over $30\,000$ offline steps. Standard SAC oscillates and ends well below the online best ($0.57/0.73$), while the anchored Bellman--actor objective rises steadily and peaks at $1.00/1.00$ around $15\,000$ steps on both source and shifted lighting.}
    \label{fig:loss-curve}
\end{figure}

We first establish the loss-side advantage and dissect the two anchor terms on USB insertion at $\alpha{=}0.75$. Figure~\ref{fig:loss-curve} compares the full anchored Bellman--actor objective against the standard SAC fine-tune objective along the training horizon: with anchoring the policy converges stably to near-perfect success on both source and shifted lighting; without it, the run exhibits the late-training collapse predicted by our drift analysis (Section~\ref{sec:method-loss}). The pattern matches our hypothesis at population scale: the anchored objective is a strictly better fine-tune objective in this regime, and the gap to the standard SAC objective widens monotonically after the joint best step around $15\,000$ iterations.

\begin{table}[H]
    \centering
    \footnotesize
    \setlength{\tabcolsep}{8pt}
    \caption{Component ablation over the two anchor terms on USB insertion at $\alpha{=}0.75$. ``og''/``re'' denote source/shifted-light evaluation; SR is success rate over 30 trajectories; ``time'' is mean successful-episode duration in seconds.}
    \label{tab:loss-ablation}
    \begin{tabular}{lcccc}
        \toprule
        \textbf{Variant} & \textbf{og SR} & \textbf{og T} & \textbf{re SR} & \textbf{re T} \\
        \midrule
        no anchor                 & 0.77 & 5.30 & 0.67 & 3.95 \\
        $\Lfeat$ only             & 0.57 & 7.69 & 0.97 & 3.45 \\
        ref.-action $\Lmse$ only  & 1.00 & 3.05 & 0.80 & 6.16 \\
        $\Lfeat + \Lmse$ (\OURS)  & \textbf{1.00} & \textbf{2.75} & \textbf{1.00} & \textbf{2.48} \\
        \bottomrule
    \end{tabular}
    \vspace{-0.6\baselineskip}
\end{table}

Table~\ref{tab:loss-ablation} further dissects the two anchor terms. The encoder-level anchor $\Lfeat$ alone substantially boosts shifted-light success ($0.67{\to}0.97$) but leaves the source-workstation undertrained on $Q$-values ($0.77{\to}0.57$); the reference-action mean anchor $\Lmse$ alone restores source-workstation success to saturation ($1.00$) but is insufficient on shifted lighting ($0.80$). Combined, the two anchors yield $1.00/1.00$ success and the shortest episode duration, confirming that they target distinct failure paths and are complementary by design.

\FloatBarrier

\subsection{\texorpdfstring{\IRR}{IRR} study: optimal original-vs.-relit mixing}
\label{sec:exp-buffer-sweep}

\begin{figure}[t]
    \centering
    \includegraphics[width=0.78\linewidth]{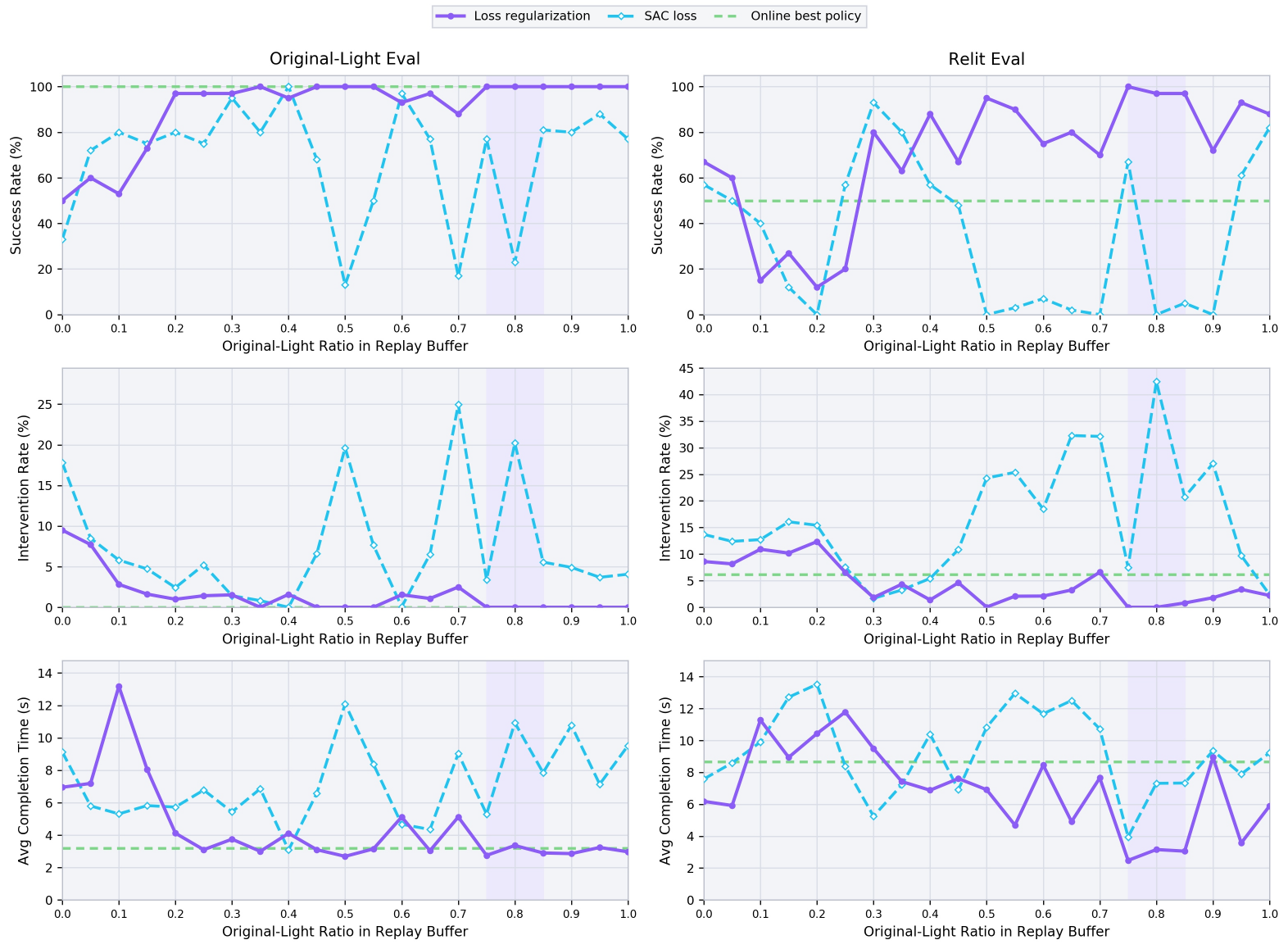}
    \caption{Sweep of the original-light fraction $\alpha$ in \IRR on USB insertion. Rows show source/shifted lighting; columns show success, intervention rate, and episode length. The anchored objective (purple) dominates standard SAC (blue) and peaks at $\alpha{=}0.75$.}
    \label{fig:alpha-sweep}
\end{figure}

We next determine the optimal original-light fraction $\alpha$ inside \IRR, sweeping $\alpha\in\{0.0, 0.05, \ldots,\allowbreak 0.95, 1.0\}$ on USB insertion. The sweep also serves as a fine-grained, in-domain check that the anchored Bellman--actor objective is preferable to the standard SAC loss across the full ratio spectrum.

Figure~\ref{fig:alpha-sweep} surfaces two findings simultaneously. First, the anchored Bellman--actor objective dominates standard SAC at essentially every value of $\alpha$ on all three metrics. Second, holding the loss fixed, the joint optimum is $\alpha{=}0.75$, where both source and shifted-light success reach $1.00$ and the average successful-episode duration drops to $2.48$~s on shifted lighting; below this point the critic loses source-workstation Bellman coverage, while at $\alpha\!\to\!1.0$ relit data vanishes from $\mathcal{R}_\alpha$ and adaptation degrades. We fix $\alpha=0.75$ in subsequent experiments.

\FloatBarrier

\subsection{Component ablation: \texorpdfstring{\IRR}{IRR} \texorpdfstring{$\times$}{x} anchored objective}
\label{sec:exp-axis-ablation}

A $2{\times}2$ design crossing ``\IRR on/off'' with ``anchored objective on/off'' (Table~\ref{tab:final-abcd}) shows that each component moves the needle but neither alone closes the gap: anchoring alone (Final-B) lifts success to $0.50/0.67$ by stabilising the encoder; \IRR alone (Final-C) keeps source-workstation success but the unconstrained encoder still drifts on shifted lighting. Only Final-D combining both reaches $1.00/1.00$, validating that data-level and representation/policy-level retention are coupled by design.

\begin{table}[!htbp]
    \centering
    \footnotesize
    \setlength{\tabcolsep}{6pt}
    \caption{$2\!\times\!2$ ablation crossing \IRR ($\alpha{=}0.75$) with the anchored Bellman--actor objective on USB insertion. \IRR ``off'' corresponds to $\alpha{=}0$.}
    \label{tab:final-abcd}
    \begin{tabular}{lcccccc}
        \toprule
        \textbf{Configuration} & \textbf{\IRR} & \textbf{Anch.\ obj.} & \textbf{og SR} & \textbf{og time} & \textbf{re SR} & \textbf{re time} \\
        \midrule
        Final-A (RLPD baseline)        & $\times$     & $\times$     & 0.33 & 9.13 & 0.57 & 7.58 \\
        Final-B (anchored obj. only)   & $\times$     & $\checkmark$ & 0.50 & 6.95 & 0.67 & 6.18 \\
        Final-C (\IRR only)            & $\checkmark$ & $\times$     & 0.77 & 5.30 & 0.67 & 3.95 \\
        Final-D (\OURS, full system)   & $\checkmark$ & $\checkmark$ & \textbf{1.00} & \textbf{2.75} & \textbf{1.00} & \textbf{2.48} \\
        \bottomrule
    \end{tabular}
\end{table}

\FloatBarrier

\subsection{Cross-method comparison}
\label{sec:exp-cross-method}

We benchmark \OURS against the five HIL-compatible baselines under controlled illumination drift on all four tasks. Table~\ref{tab:cross-method} reports the headline at $60\%$ illumination shift.

\begin{table}[!htbp]
    \centering
    \footnotesize
    \setlength{\tabcolsep}{3pt}
    \caption{Cross-method comparison at $60\%$ illumination shift ($60$ trajectories per cell). SR is success rate; ``T'' is mean successful-episode duration (s). Best per task in bold.}
    \label{tab:cross-method}
    \begin{tabular}{lcccccccc}
        \toprule
        & \multicolumn{2}{c}{\texttt{ram}} & \multicolumn{2}{c}{\texttt{usb}} & \multicolumn{2}{c}{\texttt{wipe}} & \multicolumn{2}{c}{\texttt{breaker}} \\
        \cmidrule(lr){2-3}\cmidrule(lr){4-5}\cmidrule(lr){6-7}\cmidrule(lr){8-9}
        \textbf{Method} & SR & T & SR & T & SR & T & SR & T \\
        \midrule
        BC                                  & 0.07 & 4.99  & 0.27 & 7.11 & 0.13 & 11.31 & 0.33 & 6.64 \\
        ACT~\citep{zhao2023act}             & 0.30 & 6.23  & 0.38 & 3.52 & 0.23 & 14.75 & 0.40 & 14.19 \\
        HG-DAgger~\citep{kelly2019hgdagger} & 0.13 & 8.54  & 0.37 & 6.66 & 0.17 & ---   & 0.23 & 8.54 \\
        HIL-SERL~\citep{luo2025hilserl}     & 0.43 & 5.66  & 0.50 & 4.14 & 0.33 & 13.74 & 0.13 & 6.32 \\
        IBRL~\citep{hu2024ibrl}             & 0.37 & 6.31  & 0.87 & \textbf{3.53} & 0.43 & 13.66 & 0.37 & 6.30 \\
        \textbf{\OURS} (ours)               & \textbf{1.00} & \textbf{4.54} & \textbf{1.00} & 3.59 & \textbf{0.87} & \textbf{14.39} & \textbf{0.83} & \textbf{5.67} \\
        \bottomrule
    \end{tabular}
\end{table}

\begin{figure}[!htbp]
    \centering
    \includegraphics[width=\linewidth]{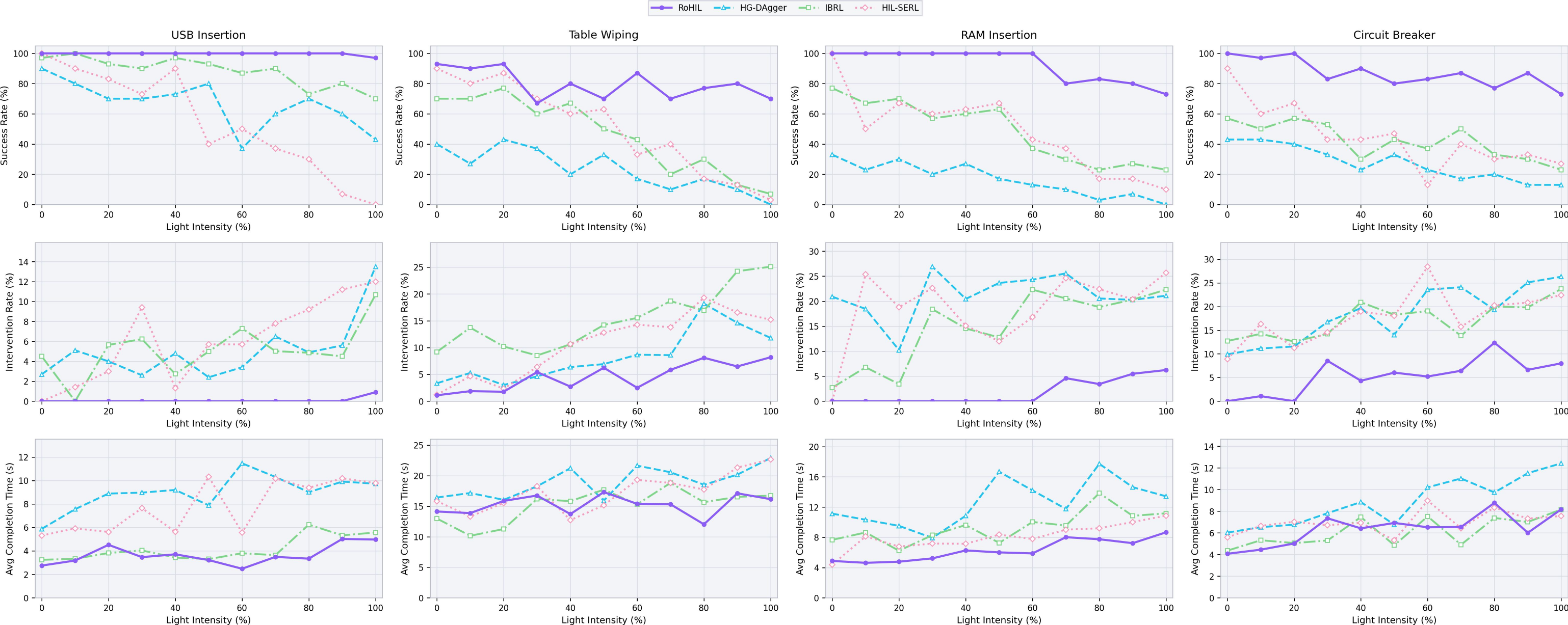}
    \caption{Lighting-shift gradient on all four tasks (rows) for \OURS vs.\ HIL-SERL, HG-DAgger, IBRL. Three metrics per row: success rate, intervention rate, mean episode length. Horizontal axis: shift intensity $\{0,10,\ldots,100\}\%$. \OURS remains essentially flat across the gradient, while baselines degrade roughly monotonically.}
    \label{fig:gradient-sweep}
\end{figure}

\OURS reaches the highest success rate on every task at $60\%$ shift (RAM $1.00$, USB $1.00$, wiping $0.87$, breaker $0.83$), while every baseline degrades materially: HIL-SERL retains barely half its source success on USB and collapses to $0.13$ on the breaker, HG-DAgger and BC fall below $0.4$ on every task, and the strongest baseline IBRL trails \OURS by $0.4$ to $0.6$ on three of four tasks. Figure~\ref{fig:gradient-sweep} extends this comparison to the full intensity gradient: \OURS stays near $1.00$ on USB throughout and falls only modestly on the other tasks at extreme shifts, while baselines degrade roughly monotonically; even IBRL loses $\geq 0.4$ between $0\%$ and $60\%$ on wiping and breaker. The gap between \OURS and the strongest baseline is consistent across tasks, indicating that the robustness is a property of the framework rather than task-specific tuning.

\section{Conclusion and Limitations}
\label{sec:conclusion}

\OURS closes the cross-workstation illumination-robustness gap of HIL-RL via three coupled mechanisms (world-model relighting, \IRR, anchored Bellman--actor regularisation), without re-collecting demonstrations and without further real-robot interaction. Three limitations point to future work: any future relighter with stronger inter-frame consistency or higher photometric fidelity should immediately translate into better shifted-light performance, since the relighter is a drop-in component in our pipeline; an anchor-free variant is needed when source-workstation demonstrations are not yet available; and extending the anchored-regularisation principle to mixed visual and geometric drift is a natural next step.

\bibliographystyle{plainnat}
\bibliography{references}
\clearpage

\appendix
\section{Algorithm and Actor-Anchor Variant}
\label{app:algorithm}

\paragraph{Pseudocode.}
Algorithm~\ref{alg:rohil} gives one Learner step of the offline fine-tune.

\begin{algorithm}[H]
\caption{\OURS offline fine-tune (one Learner step)}
\label{alg:rohil}
\begin{algorithmic}[1]
\Require frozen source-workstation parameters $\theta_0$; current $\theta$; target $\bar\theta$; \IRR replay $\mathcal{R}_\alpha$ with $\alpha{=}0.75$; anchor pool $\mathcal{D}_{\mathrm{anc}}$ (original-light human demos $\cup$ relit demos); weights $\lambda_{\mathrm{feat}}, \beta_{\mathrm{mse}}$; schedule $\rho(t)$
\State Sample $\mathcal{B}_R\!\sim\!\mathcal{R}_\alpha$ ($|\mathcal{B}_R|{=}B/2$), $\mathcal{B}_D\!\sim\!\mathcal{D}_{\mathrm{anc}}$ ($|\mathcal{B}_D|{=}B/2$); form $\mathcal{B}=\mathcal{B}_R\cup\mathcal{B}_D$
\State Compute SAC target $y = r + \gamma(1-d)\left[\min_i Q_{\bar\theta_i}(s',a')-\eta\log\pi_\theta(a'|s')\right]$, $a'\!\sim\!\pi_\theta(\cdot|s')$
\State $\Lbellman \gets \mathbb{E}_\mathcal{B}\bigl[\sum_i(Q_{\theta_i}(s,a)-\sg(y))^2\bigr]$
\State $\Lfeat \gets \lambda_{\mathrm{feat}}\,\rho(t)\,\mathbb{E}_{\mathcal{B}_D}\bigl[\|\phi_\theta(o)-\sg(\phi_{\theta_0}(o))\|_F^2\bigr]$
\State Update critic on $\Lcritic \!=\! \Lbellman + \Lfeat$
\State $\Lsac \gets \mathbb{E}_{s\sim\mathcal{B},\,a\sim\pi_\theta(\cdot|s)}\bigl[\eta\log\pi_\theta(a|s)-\min_i Q_{\theta_i}(s,a)\bigr]$
\State $\Lmse  \gets \beta_{\mathrm{mse}}\,\rho(t)\,\mathbb{E}_{\mathcal{B}_D}\bigl[\|\mu_\theta(s)-\sg(\mu_{\theta_0}(s))\|_2^2\bigr]$
\State Update actor on $\Lactor  \!=\! \Lsac     + \Lmse$
\State Polyak-update target $\bar\theta \!\leftarrow\! \tau\theta + (1\!-\!\tau)\bar\theta$
\end{algorithmic}
\end{algorithm}

\paragraph{Actor anchor head: reference-action mean vs.\ KL.}
\label{app:actor-head}
The policy anchor admits two natural forms on the SAC diagonal-Gaussian: our reference-action mean anchor $\Lmse=\|\mu_\theta-\sg(\mu_{\theta_0})\|_2^2$, used inside $\Lactor$ in the main method (Equation~\ref{eq:lmse}); and a full distributional KL $\Lkl=D_{\mathrm{KL}}(\pi_{\theta_0}\|\pi_\theta)$, closed-form on diagonal-Gaussians, which constrains both the mean and the variance. KL is the stronger anchor but in our setting we want stability without throttling adaptation. Empirically (Table~\ref{tab:actor-head}), the two heads tie on the source workstation, but KL slightly under-performs the reference-action mean anchor on shifted lighting ($0.97$ vs.\ $1.00$ success, $3.27$~s vs.\ $2.48$~s mean episode duration), consistent with the hypothesis that locking the variance throttles relit-domain adaptation.

\begin{table}[!htbp]
\centering
\footnotesize
\setlength{\tabcolsep}{5pt}
\caption{Actor anchor head: reference-action mean anchor (ours) vs.\ full KL distillation on USB insertion at $\alpha{=}0.75$. ``og''/``re'' denote source/shifted-light evaluation; SR over $30$ trajectories.}
\label{tab:actor-head}
\begin{tabular}{lcccccc}
\toprule
\textbf{Anchor head} & \textbf{og SR} & \textbf{og time} & \textbf{og interv.} & \textbf{re SR} & \textbf{re time} & \textbf{re interv.} \\
\midrule
$\Lmse$ on $\mu$ (ours)  & 1.00 & 2.75 & 0.00\% & \textbf{1.00} & \textbf{2.48} & \textbf{0.00\%} \\
$\Lkl$ on $(\mu,\sigma)$ & 1.00 & \textbf{2.46} & 0.00\% & 0.97 & 3.27 & 1.00\% \\
\bottomrule
\end{tabular}
\end{table}

\FloatBarrier

\section{Per-Task Setup and Hyperparameters}
\label{app:tasks}

\begin{figure}[!htbp]
    \centering
    \includegraphics[width=0.9\linewidth]{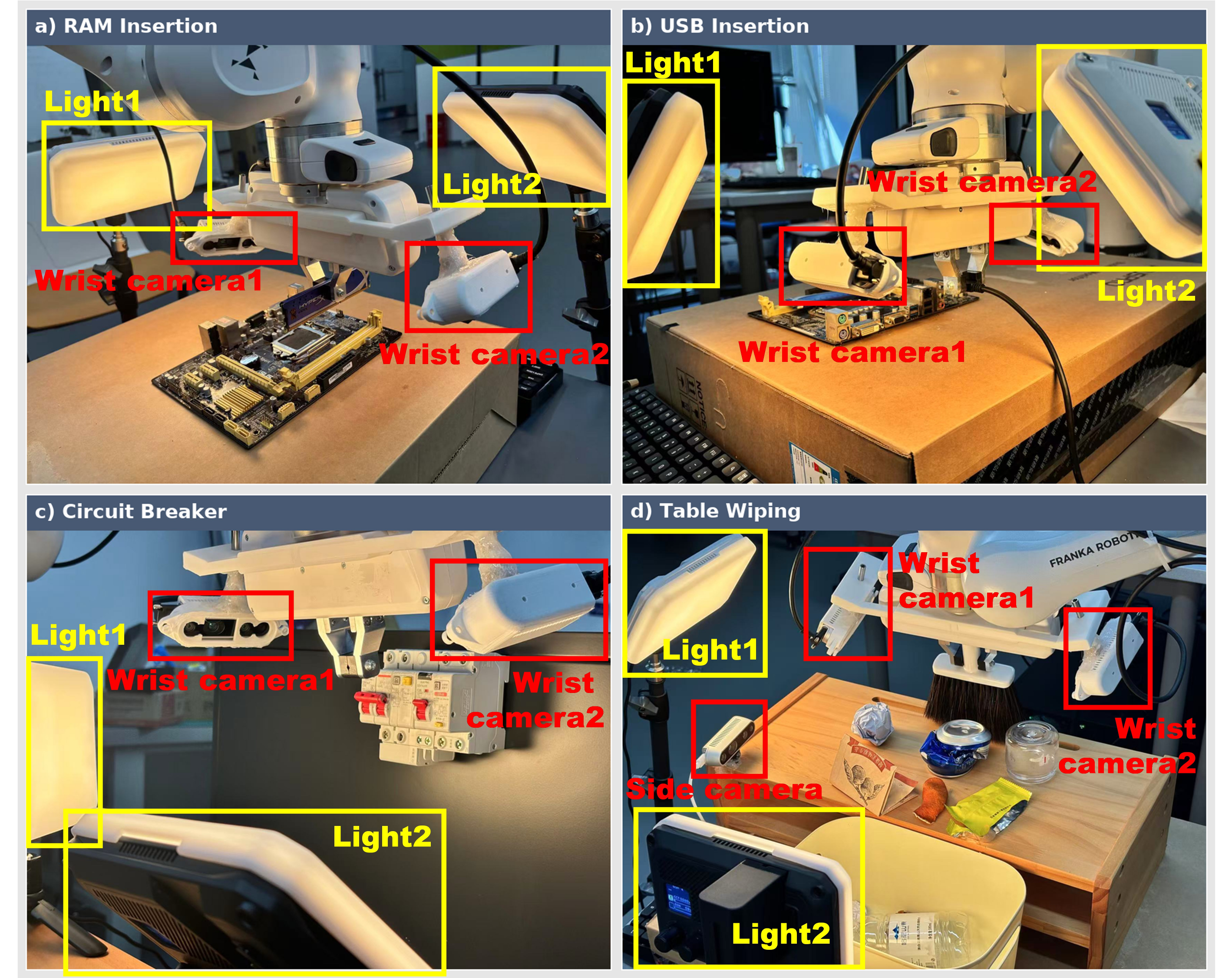}
    \caption{Hardware mounts and camera placements for the four tasks.}
    \label{fig:hardware}
\end{figure}

The four tasks share the Franka Emika Panda arm, a wrist-mounted Intel RealSense camera, and the SpaceMouse intervention channel; they differ in initial demonstration count, episode length, and reset procedure. The action space is uniformly 6-dimensional and corresponds to the end-effector delta twist. Across all tasks the shared training hyperparameters are $\lambda_{\mathrm{feat}}=0.2$, $\beta_{\mathrm{mse}}=0.1$, $\rho_{\mathrm{end}}=0.33$, and original-light fraction $\alpha=0.75$. Figure~\ref{fig:hardware} summarises the corresponding hardware mounts and camera placements.

\paragraph{a) RAM insertion.} The robot grasps a fixed RAM module and inserts it into a vertical motherboard slot, requiring sub-millimetre alignment of the gold contact strip with the slot teeth. We report the policy training details for this task in Table~\ref{tab:hp-ram}.

\begin{table}[H]
\centering
\footnotesize
\setlength{\tabcolsep}{4pt}
\caption{Policy training details for the RAM insertion task.}
\label{tab:hp-ram}
\begin{tabular}{@{}p{0.34\linewidth}p{0.60\linewidth}@{}}
\toprule
\textbf{Parameter} & \textbf{Value} \\
\midrule
Action space                 & 6-dimensional end-effector delta twist \\
Initial demonstrations       & 20 \\
Max episode length           & 150 steps \\
Reset method                 & Script reset \\
Source policy training       & $60\,000$ HIL-SERL steps \\
Offline robust fine-tune     & $15\,000$ learner-only steps; no additional robot interaction \\
Fine-tune batch size         & 256 transitions per learner step \\
Policy / critic backbone     & SAC + RLPD replay, ResNet-10 encoder \\
Relighting expansion         & DiffusionRenderer; 4 relit + 1 original \\
\IRR replay                  & $\mathcal{B}=\frac{1}{2}\mathrm{Sample}(\mathcal{R}_\alpha)\cup\frac{1}{2}\mathrm{Sample}(\mathcal{D}_{\mathrm{anc}})$, $\alpha=0.75$ \\
Anchor pool                  & Original-light demos $\cup$ relit demos \\
Anchor weights / schedule    & $\lambda_{\mathrm{feat}}=0.2$, $\beta_{\mathrm{mse}}=0.1$, $\rho_{\mathrm{end}}=0.33$ \\
\bottomrule
\end{tabular}
\end{table}

\paragraph{b) USB insertion.} The robot inserts a USB connector into a fixed receptacle. The small contact patch and the polarised orientation make this the most lighting-sensitive of the four tasks: the gripper-jaw glare under shifted lighting is what fails the source policy. We report the policy training details for this task in Table~\ref{tab:hp-usb}.

\begin{table}[H]
\centering
\footnotesize
\setlength{\tabcolsep}{4pt}
\caption{Policy training details for the USB insertion task.}
\label{tab:hp-usb}
\begin{tabular}{@{}p{0.34\linewidth}p{0.60\linewidth}@{}}
\toprule
\textbf{Parameter} & \textbf{Value} \\
\midrule
Action space                 & 6-dimensional end-effector delta twist \\
Initial demonstrations       & 20 \\
Max episode length           & 150 steps \\
Reset method                 & Script reset \\
Source policy training       & $30\,000$ HIL-SERL steps \\
Offline robust fine-tune     & $15\,000$ learner-only steps; no additional robot interaction \\
Fine-tune batch size         & 256 transitions per learner step \\
Policy / critic backbone     & SAC + RLPD replay, ResNet-10 encoder \\
Relighting expansion         & DiffusionRenderer; 4 relit + 1 original \\
\IRR replay                  & $\mathcal{B}=\frac{1}{2}\mathrm{Sample}(\mathcal{R}_\alpha)\cup\frac{1}{2}\mathrm{Sample}(\mathcal{D}_{\mathrm{anc}})$, $\alpha=0.75$ \\
Anchor pool                  & Original-light demos $\cup$ relit demos \\
Anchor weights / schedule    & $\lambda_{\mathrm{feat}}=0.2$, $\beta_{\mathrm{mse}}=0.1$, $\rho_{\mathrm{end}}=0.33$ \\
\bottomrule
\end{tabular}
\end{table}

\paragraph{c) Circuit-breaker actuation.} The robot toggles a circuit-breaker switch from off to on by hooking the switch lever and lifting it. The challenge is to control the contact force between the lever and the gripper finger so that the lever rotates without slipping. We report the policy training details for this task in Table~\ref{tab:hp-circuit}.

\begin{table}[H]
\centering
\footnotesize
\setlength{\tabcolsep}{4pt}
\caption{Policy training details for the circuit-breaker actuation task.}
\label{tab:hp-circuit}
\begin{tabular}{@{}p{0.34\linewidth}p{0.60\linewidth}@{}}
\toprule
\textbf{Parameter} & \textbf{Value} \\
\midrule
Action space                 & 6-dimensional end-effector delta twist \\
Initial demonstrations       & 20 \\
Max episode length           & 150 steps \\
Reset method                 & Human reset \\
Source policy training       & $35\,000$ HIL-SERL steps \\
Offline robust fine-tune     & $15\,000$ learner-only steps; no additional robot interaction \\
Fine-tune batch size         & 256 transitions per learner step \\
Policy / critic backbone     & SAC + RLPD replay, ResNet-10 encoder \\
Relighting expansion         & DiffusionRenderer; 4 relit + 1 original \\
\IRR replay                  & $\mathcal{B}=\frac{1}{2}\mathrm{Sample}(\mathcal{R}_\alpha)\cup\frac{1}{2}\mathrm{Sample}(\mathcal{D}_{\mathrm{anc}})$, $\alpha=0.75$ \\
Anchor pool                  & Original-light demos $\cup$ relit demos \\
Anchor weights / schedule    & $\lambda_{\mathrm{feat}}=0.2$, $\beta_{\mathrm{mse}}=0.1$, $\rho_{\mathrm{end}}=0.33$ \\
\bottomrule
\end{tabular}
\end{table}

\paragraph{d) Table wiping.} The robot wipes a circular region on a flat work surface with a cloth attached to the end-effector, applying a moderate downward force throughout. The longer episode length reflects the multi-pass nature of this task. We report the policy training details for this task in Table~\ref{tab:hp-wipe}.

\begin{table}[H]
\centering
\footnotesize
\setlength{\tabcolsep}{4pt}
\caption{Policy training details for the table wiping task.}
\label{tab:hp-wipe}
\begin{tabular}{@{}p{0.34\linewidth}p{0.60\linewidth}@{}}
\toprule
\textbf{Parameter} & \textbf{Value} \\
\midrule
Action space                 & 6-dimensional end-effector delta twist \\
Initial demonstrations       & 30 \\
Max episode length           & 250 steps \\
Reset method                 & Human reset \\
Source policy training       & $95\,000$ HIL-SERL steps \\
Offline robust fine-tune     & $15\,000$ learner-only steps; no additional robot interaction \\
Fine-tune batch size         & 256 transitions per learner step \\
Policy / critic backbone     & SAC + RLPD replay, ResNet-10 encoder \\
Relighting expansion         & DiffusionRenderer; 4 relit + 1 original \\
\IRR replay                  & $\mathcal{B}=\frac{1}{2}\mathrm{Sample}(\mathcal{R}_\alpha)\cup\frac{1}{2}\mathrm{Sample}(\mathcal{D}_{\mathrm{anc}})$, $\alpha=0.75$ \\
Anchor pool                  & Original-light demos $\cup$ relit demos \\
Anchor weights / schedule    & $\lambda_{\mathrm{feat}}=0.2$, $\beta_{\mathrm{mse}}=0.1$, $\rho_{\mathrm{end}}=0.33$ \\
\bottomrule
\end{tabular}
\end{table}

\FloatBarrier

\section{Lighting Conditions and Relighter Comparison}
\label{app:lighting-conditions}\label{app:relighting-compare}

\begin{figure}[!htbp]
    \centering
    \includegraphics[width=0.9\linewidth]{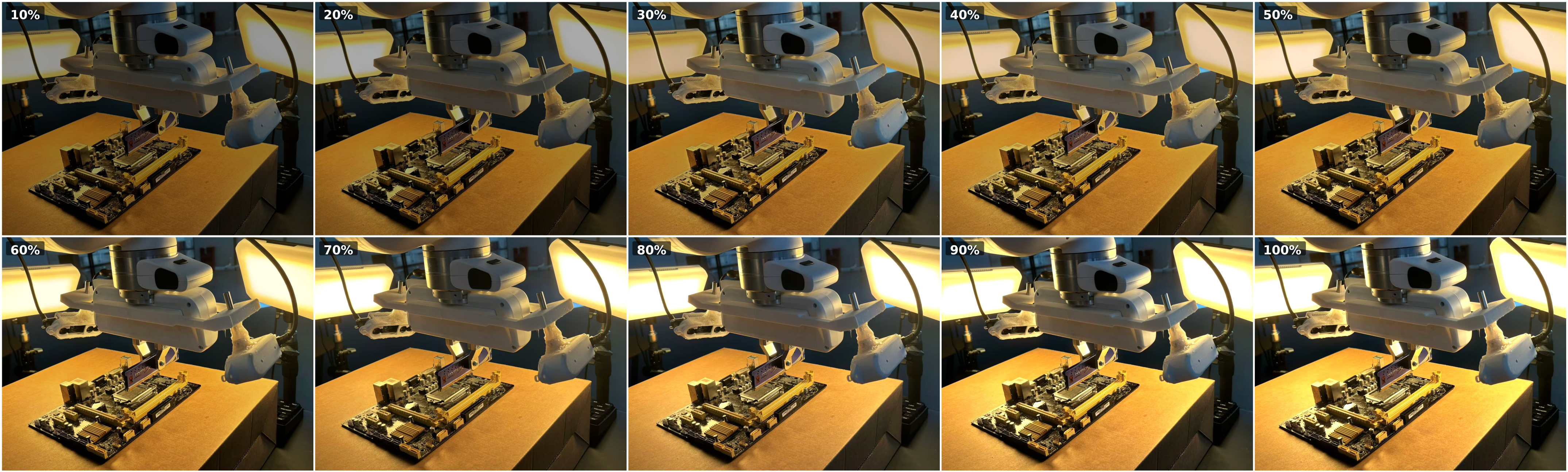}
    \caption{The ten illumination conditions used in our evaluation. Top: five HDRI environment maps. Middle: three task-light spotlight configurations. Bottom: two natural-window-light shifts.}
    \label{fig:lighting-grid}
\end{figure}

\begin{figure}[!htbp]
    \centering
    \includegraphics[width=0.9\linewidth]{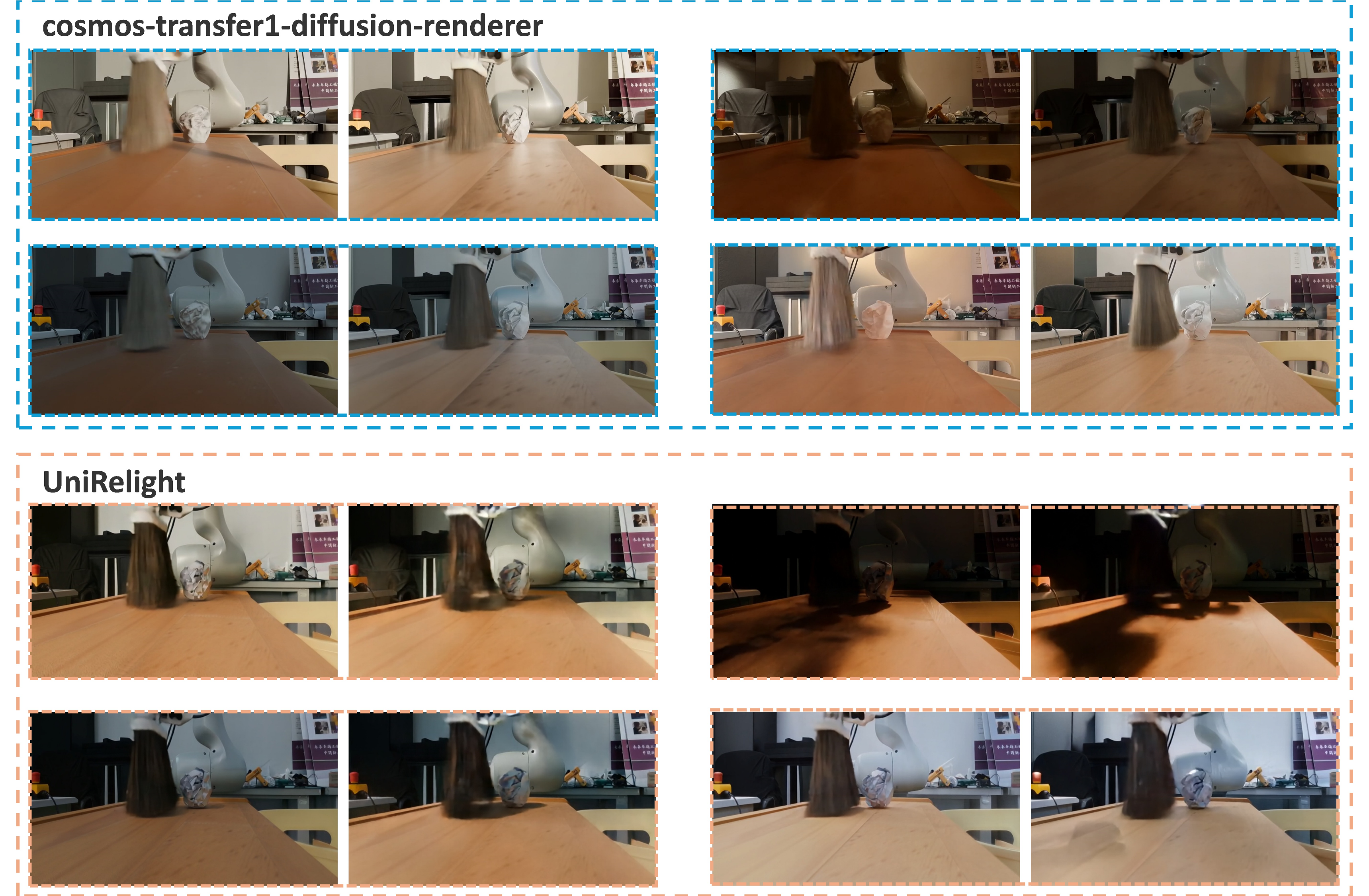}
    \caption{Visual comparison of Cosmos-Transfer1-DiffusionRenderer (top) and UniRelight (bottom) on a representative manipulation frame across four HDRI conditions. UniRelight exhibits stronger inter-frame temporal consistency; DiffusionRenderer is roughly $7\times$ cheaper to run.}
    \label{fig:relighter-quality}
\end{figure}

We report the ten illumination conditions used in our evaluation in Figure~\ref{fig:lighting-grid}, and a visual / cost comparison of the candidate relighters in Figure~\ref{fig:relighter-quality} and Table~\ref{tab:relighter-cost}.

\begin{table}[!htbp]
\centering
\footnotesize
\setlength{\tabcolsep}{5pt}
\caption{Resource cost of relighting one camera stream of $8\,000$ transitions on the same single-GPU hardware. Both methods are run at bf16 for a fair comparison; UniRelight uses a minimal inference-time loader fix that removes a transient load-time VRAM spike, while preserving the released bf16 inference setting.}
\label{tab:relighter-cost}
\begin{tabular}{llrrl}
\toprule
\textbf{Method} & \textbf{Precision} & \textbf{Runtime} & \textbf{Peak VRAM} & \textbf{Notes} \\
\midrule
DiffusionRenderer & bf16 & $6.82$~h  & $26.80$~GB & adopted in this work \\
UniRelight        & bf16 & $48.63$~h & $39.88$~GB & finer temporal continuity, $\sim 7\times$ slower \\
\bottomrule
\end{tabular}
\end{table}

\FloatBarrier

\end{document}